\documentclass[lettersize,journal]{IEEEtran}

\usepackage{amsmath,amsfonts,amssymb}
\usepackage{algorithmic}
\usepackage{algorithm}
\usepackage{array}
\usepackage[caption=false,font=normalsize,labelfont=sf,textfont=sf]{subfig}
\usepackage{textcomp}
\usepackage{stfloats}
\usepackage{url}
\usepackage{verbatim}
\usepackage{graphicx}
\usepackage{cite}
\usepackage{hyperref}
\usepackage{enumitem}

\usepackage{booktabs}
\usepackage{multirow}
\usepackage{makecell}
\usepackage{tabularx}
\usepackage{pifont}
\usepackage{bm}
\usepackage{xcolor}
\usepackage{colortbl}
\usepackage{pgfplots}
\usepackage{pgfplotstable}
\pgfplotsset{compat=1.18}
\usepackage{microtype}

\providecommand{\citep}[1]{\cite{#1}}
\providecommand{\citet}[1]{\cite{#1}}

\definecolor{lightgray}{gray}{0.92}

\begin{document}

\title{Low-Cost Black-Box Detection of LLM Hallucinations via Dynamical System Prediction}

\author{Dan~Wilson$^*$,
        and~Mohamed~Akrout$^*$,~\IEEEmembership{Member,~IEEE}%
\thanks{$^*$Equal contribution.}%
\thanks{D. Wilson and M. Akrout are with the Department of Electrical Engineering and Computer Science, University of Tennessee, Knoxville, TN 37996, USA (e-mail: dwilso81@tennessee.edu; makrout@tennessee.edu).}%
\thanks{Manuscript received September 25, 2026.}}

\markboth{}%
{Wilson and Akrout: Low-Cost Black-Box Detection of LLM Hallucinations via Dynamical System Prediction}

\IEEEpubid{0000--0000/00\$00.00~\copyright~2026 IEEE}

\maketitle

\begin{abstract}
Large Language Models (LLMs) frequently generate plausible but non-factual content, a phenomenon known as hallucination. While existing detection methods typically rely on computationally expensive sampling-based consistency checks or external knowledge retrieval, we propose a new method that treats the LLM as a black-box dynamical system. By projecting LLM responses into a high-dimensional manifold via an embedding model, we characterize the resulting vector sequences as observable realizations of the model's latent state-space dynamics. Leveraging Koopman operator theory, we fit the transition operators for both factual and hallucinated regimes and define a differential residual score based on their respective prediction errors. This approach enables low-cost hallucination detection in a single-sample pass, avoiding the need for secondary sampling or external grounding. Extensive testing across three data benchmarks demonstrates that our method achieves state-of-the-art performance with reduced resource overhead.
\end{abstract}


\section{Introduction}
\IEEEPARstart{L}{arge} Language Models (LLMs) have achieved remarkable success across diverse natural language processing tasks \citep{brown2020language, achiam2023gpt}. However, their tendency to generate fluent yet factually incorrect content, commonly known as hallucination, remains a critical barrier to reliable deployment in high-stakes domains such as healthcare, law, finance, and autonomous agents \citep{huang2025survey,warraich2025ethical}. As LLMs become increasingly integrated into real-world applications, the ability to detect and mitigate hallucinations has emerged as a pressing research priority.

Hallucination in LLMs is not merely a design flaw but a statistical necessity \cite{huang2025survey}. Any calibrated language model must inevitably produce false or unsubstantiated claims, establishing hallucination as theoretically unavoidable \cite{kalai2024calibrated}. This phenomenon emerges from the fundamental trade-off between finite training data and open-ended generation \cite{bender2021dangers}. Empirical evidence shows that even state-of-the-art models cannot reliably distinguish belief from knowledge \cite{kalai2025language,suzgun2025language,lin2022truthfulqa,lin2022truthfulqa}. Ultimately, the persistent reliance on external verification \cite{li2023benchmarking} highlights a critical gap in LLM autonomy. To achieve scalable trust, it is critical to develop self-contained detection frameworks that identify hallucinations with minimal to no reliance on external knowledge retrieval.

Hallucination detection methods are generally categorized by their level of model access. \textit{White-box} approaches leverage internal representations like hidden states or attention maps \citep{azaria2023internal, su2024unsupervised, chen2024inside}, while \textit{gray-box} methods utilize token-level output probabilities \cite{qian2025beyond, bar2025learning}. However, both remain impractical for many commercial LLMs accessed via restricted application programming interfaces (APIs). \textit{Black-box} methods circumvent these access barriers by operating solely on generated text, typically through knowledge retrieval or sampling-based consistency checks, such as \texttt{SelfCheckGPT} \citep{manakul2023selfcheckgpt}. Existing black-box approaches suffer from significant computational overhead: retrieval-based methods depend on external knowledge latency, and sampling-based methods require multiple forward passes. This creates a dual barrier of high API costs and latency for real-time applications.

\IEEEpubidadjcol

\begin{figure*}[!t]
    \centering
    \includegraphics[scale=0.35]{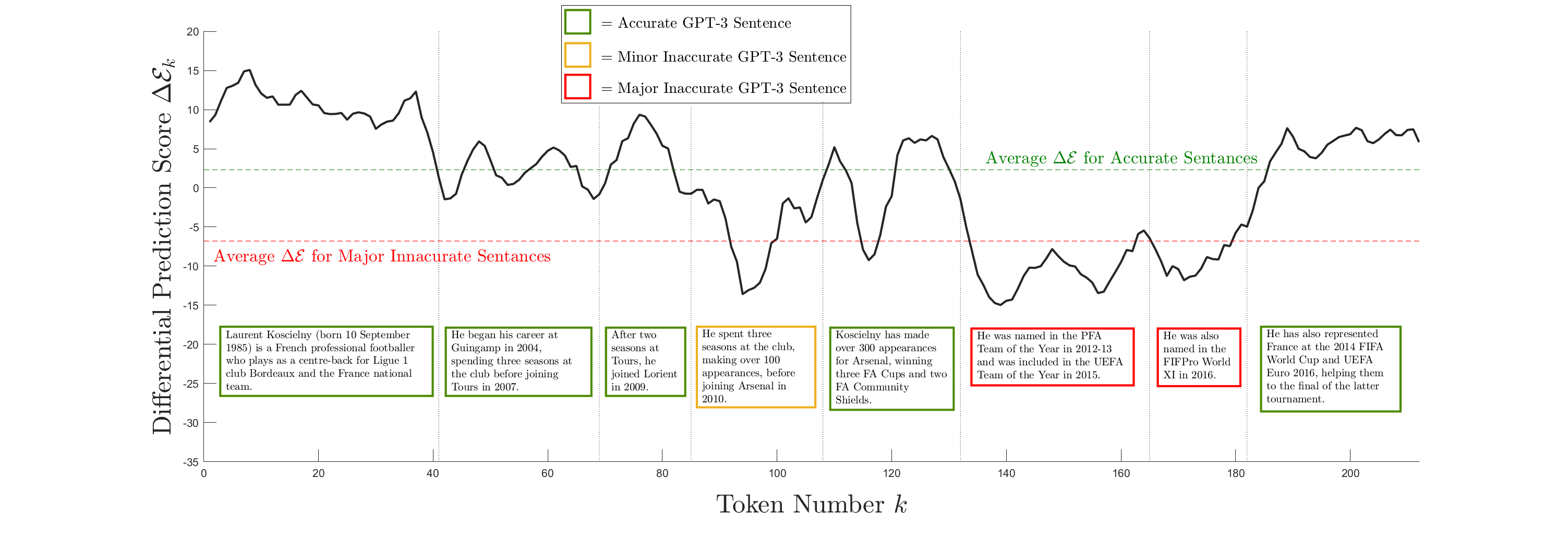} 
    \vspace{-0.3cm}
    \caption{Our proposed differential residual score $\Delta\mathcal{E}$ accurately captures the transition between factual and hallucinated sentences in segmented LLM responses from the \texttt{WikiBio} dataset. The score is computed by comparing the relative accuracy between predictions of the dynamics of token embeddings from \texttt{Llama-3} with dynamical system models inferred from both hallucinated and non-hallucinated responses.}
    \label{fig:example-sentence-embedding}
    \vspace{-0.2cm}
\end{figure*}

To address these limitations, we treat the LLM as dynamical system (DS) and propose a novel DS prediction approach that considers the temporal evolution of generated token embeddings. Our method identifies hallucinations by classifying the evolution of the output token embeddings by comparing the relative accuracy of predictive models trained from both hallucinated and non-hallucinated responses allowing for the definition of a differential prediction score $\Delta\mathcal{E}$. Fig.~\ref{fig:example-sentence-embedding} provides an example of this strategy showing the evolution of this score over the token embedding of an LLM response. Here $\Delta\mathcal{E}$ rises and falls with the factuality of LLM's output sentences in the embedding space. We build upon the differential prediction score to provide a hallucination detection technique that uses one-single pass detection without the need for output token probabilities or access to external information. To our knowledge, this is the first work to consider the output of an LLM as a black-box dynamical system with hallucinations and factual responses stemming from trajectories evolving on different manifolds embedded in the state space. Our primary contributions are as follows:
\begin{enumerate}[leftmargin=*]
\item We devise a new method for hallucination detection that treats LLM generation as a dynamical system with a high-dimensional observable space. Results suggest that the temporal evolution of the observables stemming from factual and hallucinated responses evolve on distinct manifolds that can be distinguished from one another.
\item We introduce a computationally efficient hallucination detection algorithm using a differential error score between the prediction for trajectories evolving on the accurate and inaccurate manifolds. Unlike traditional black-box metrics that necessitate multiple samples or external grounding, our method identifies hallucinations using a one-sample pass. This is achieved by fitting linear dynamical systems on a small set of LLM responses, bypassing the need for computationally heavy retrieval modules.
\item We establish the state-of-the-art performance of our DS method through an extensive evaluation across three data benchmarks: \texttt{FELM} \cite{zhao2023felm}, \texttt{HaluEval} \cite{li2023halueval}, and \texttt{WikiBio} \cite{manakul2023selfcheckgpt}. We benchmark our approach against established methods that generally require significantly greater resources, including multi-sample consistency checks and/or external knowledge retrieval. Despite the absence of comparable single-pass retrieval-free black-box baselines in current literature, our DS approach consistently outperforms or matches these resource-intensive benchmarks.
\end{enumerate}

\section{Related Work}\label{sec:related-work}
Hallucination detection methods differ in the degree of model access they require, and this difference affects their applicability across different deployment settings (see \cite{huang2025survey} for a comprehensive review).

\textbf{White-box and gray-box methods.} White-box methods require full access to the internal representations of the model, including hidden-state activations, embedding layers, attention maps, and gradients \cite{azaria2023internal,chen2024inside, su2024unsupervised, hu2024embedding}. By directly inspecting these internal signals, white-box approaches can capture fine-grained indicators of hallucination that are invisible at the output level \cite{sriramanan2024llm}. Despite their strong performance, white-box methods are typically feasible only when the model is hosted locally, as they require unrestricted access to the model architecture and its intermediate computations. Gray-box methods, however, relax the access requirements of white-box approaches by leveraging the output probability distribution over tokens, including next-token probabilities \cite{quevedo2024detecting} and logit entropy \cite{farquhar2024detecting,DBLP:conf/iclr/KuhnGF23} which are commonly exposed by commercial LLM APIs without granting access to internal model states \cite{qian2025beyond,bar2025learning,qiao2026lowest}.

\textbf{Black-box methods.} Black-box methods impose the fewest access requirements because they operate exclusively on the generated output tokens with no visibility into internal states or probability distributions. These approaches typically rely on sampling multiple responses and evaluating their consistency through lexical overlap, entailment-based comparison, or knowledge-graph representations of factual claims \cite{manakul2023selfcheckgpt, sawczyn2026factselfcheck, zhang2023sac3, kong2025multiperspective,goel2025zero}. For example, Manakul et al.\ propose \texttt{SelfCheckGPT}, which detects hallucinations by measuring the consistency among multiple stochastically sampled responses under the assumption that factual statements tend to be reproduced consistently, whereas hallucinated content varies across samples \cite{manakul2023selfcheckgpt}. Zhang et al.\ introduce SAC, which performs semantic-aware cross-check consistency verification to identify unreliable outputs \cite{zhang2023sac3}. Kong et al.\ further extend this line of work by proposing a multi-perspective consistency checking framework that operates in a zero-resource setting, requiring neither external knowledge bases nor access to model internals \cite{kong2025multiperspective}. Because black-box methods depend solely on the generated text, they offer the broadest deployability and are compatible with any LLM, including fully closed-source systems \cite{farquhar2024detecting}.

\section{Method}
\begin{figure}[!b]
\vspace{-1cm}
\centering
    \subfloat[\footnotesize{\texttt{Qwen3} Modeshapes}]{{
    \includegraphics[width=0.46\columnwidth]{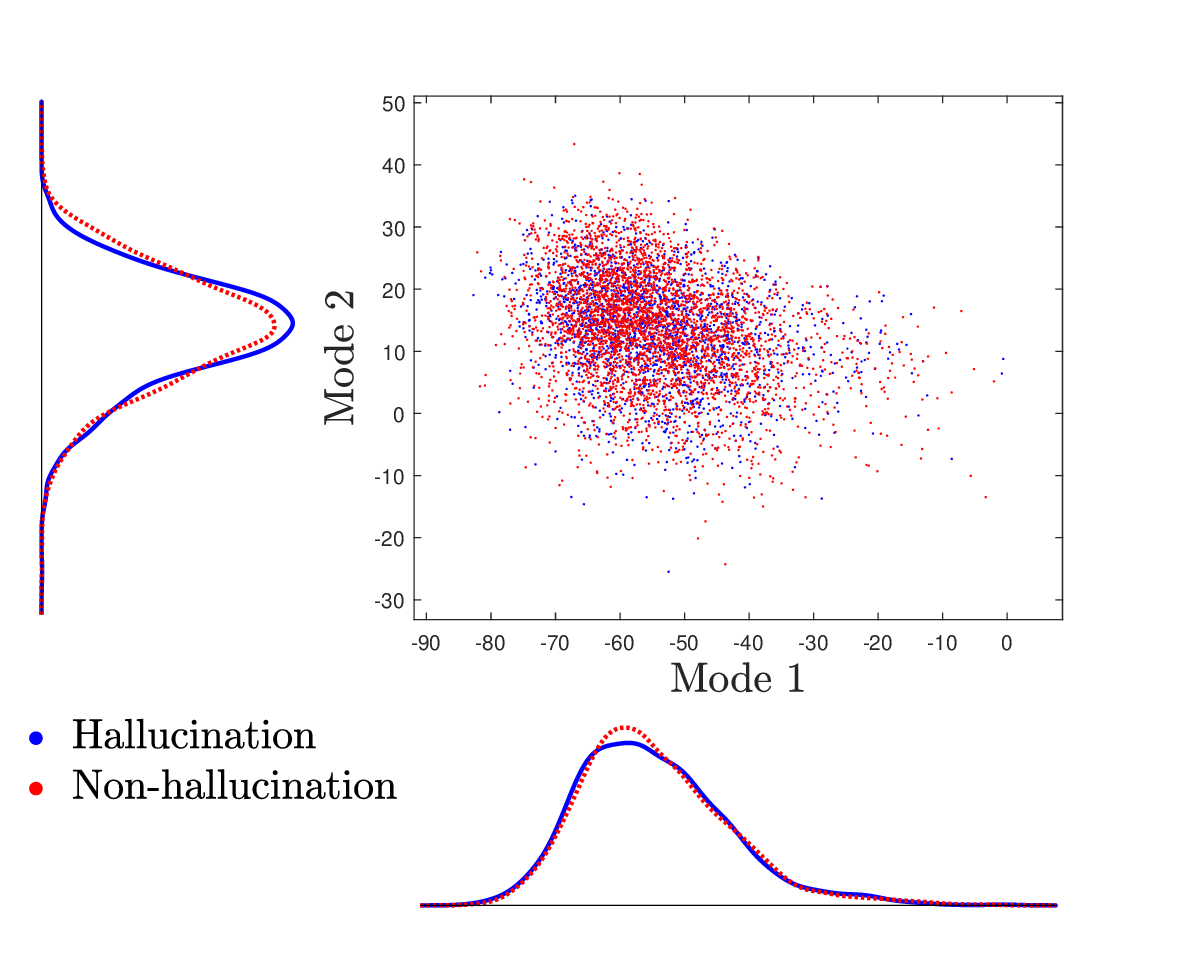}
    \label{fig:modeshape-qwen}}}
    \hfill
    \subfloat[\footnotesize{\texttt{Llama-3} Modeshapes}]{{
    \includegraphics[width=0.46\columnwidth]{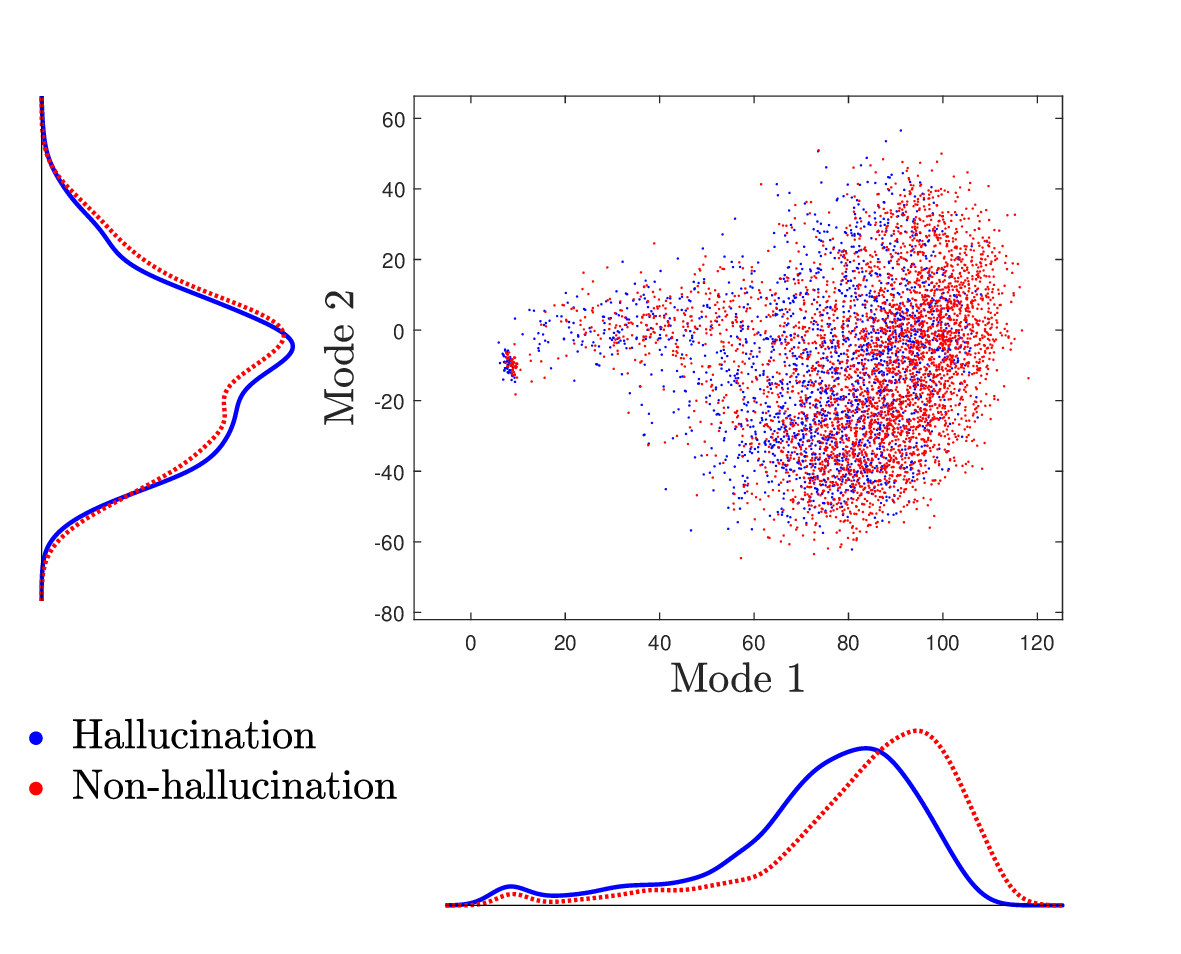}
    \label{fig:modeshape-llama}}}
    \caption{For 300 factual ground-truth and hallucinated responses from the \texttt{HaluEval} dataset, the magnitudes of the dominant SVD modes from \texttt{Qwen} and \texttt{Llama-3} token embeddings are shown in (a) and (b), respectively. While the probability distributions show minor differences between hallucinated and factual responses, these differences are subtle and would require a prohibitively long token sequence for accurate sample classification.}
    \label{fig:modeshapes}
\end{figure}

Moving from traditional token-level analysis to a dynamical systems perspective, we address the limitations of treating the output of LLM outputs as a sequence of independent tokens. Existing black-box methods rely on lexical overlap or semantic consistency between multiple sampled responses \cite{goel2025zero,zhang2023sac3}. However, as illustrated in Fig.~\ref{fig:modeshapes}, the first two embedding singular value decomposition (SVD) modes of correct and hallucinated samples are not sufficient to distinguish between factual and hallucinated responses from the LLM and are dependent on the embedding model. To design a hallucination detection algorithm, we present a method to probe the dynamics of the token embeddings.

\subsection{Overview}
We consider an LLM and its output as a dynamical system of the form:
\begin{align} \label{dynsyst}
    \mathbf{x}_{k+1} &= F(\mathbf{x}_{k}), \nonumber \\
    \mathbf{q}_k &= G_1(\mathbf{x}_k), \nonumber \\
    \mathbf{y}_k &= H(\mathbf{x}_{k}) = G_2(G_1(\mathbf{x}_k)).
\end{align}

\begin{table}[!t]
\caption{Conceptual mapping between dynamical systems and LLM variables.}
\label{table:analogy-LLM}
\centering
\begin{tabular}{@{} c p{5.5cm} @{}}
    \toprule
    \textbf{Variable} & \textbf{Corresponding Element of LLMs} \\
    \midrule
    $\mathbf{x}_k$ & Full set of activations maps \\
    $F$ & Transformer's forward operations \\
    $G_1$ & Softmax and sampling \\
    $G_2$ & Embedding model \\
    \bottomrule
\end{tabular}
\vspace{-0.5cm}
\end{table}

In \eqref{dynsyst}, $\mathbf{x}_{k} \in \mathbb{R}^N$ represents state variables internal to the LLM immediately before the next token $\mathbf{q}_k$ is chosen, and $F$ is a nonlinear map that determines the state evolution between tokens. Here, $\mathbf{q}_k$ is in the space of all possible tokens produced by the LLM, i.e.,~its vocabulary. We take $G_1$ to be a nonlinear function that maps the state to a token and $G_2$ to be map between the token and the embedding (via \texttt{Qwen}, \texttt{Mistral}, etc.). The dimension $N$ is on the order of millions or billions depending on the specific LLM used and is generally not accessible. We take $\mathbf{y}_k \in \mathbb{R}^M$ to be the observable where $M$ depends on the embedding model used. Table~\ref{table:analogy-LLM} summarizes the conceptual mapping between the dynamical system from \eqref{dynsyst} and the LLM's internal modules. We treat the LLM as a discrete-time non-linear black-box system and employ Koopman operator theory as a foundation for our approach to characterize the dynamics of observables. From this perspective, the Koopman operator $K: \mathbb{R}^M \rightarrow \mathbb{R}^M$ is defined as
\begin{equation} \label{kaut}
    K H(\mathbf{x}_k) \equiv H(F(\mathbf{x}_k)).
\end{equation}
Despite the fact that the functions $F$ and $H$ are nonlinear, the Koopman operator is linear (a property inherited from the linearity of the composition \cite{budi12, mezi13}) but generally infinite-dimensional.

To build a model for classification, we use LLM output data that consists of both hallucinations and correct responses in order to infer Koopman-based dynamical models for each. From a dynamics systems perspective, we posit that correct and hallucinated responses from the LLM represent solutions evolving on different manifolds $\mathcal{M}_h$, $\mathcal{M}_c \in \mathbb{R}^N$, respectively so that Koopman-based model identification strategies applied to hallucinated and correct datasets will yield different estimates for the action of the Koopman operator \cite{budi12}. Models that were inferred on data from hallucinated (resp.,~non-hallucinated) responses will better predict the dynamics of the token embedding for hallucinated (resp.,~non-hallucinated) responses. The relative accuracy between these estimates can subsequently be used for classification on a sentence-level and response-level basis.

\begin{figure*}[!t]
    \centering
    \subfloat[Phase 1: Dynamical system fitting]{{
    \includegraphics[width=0.37\textwidth]{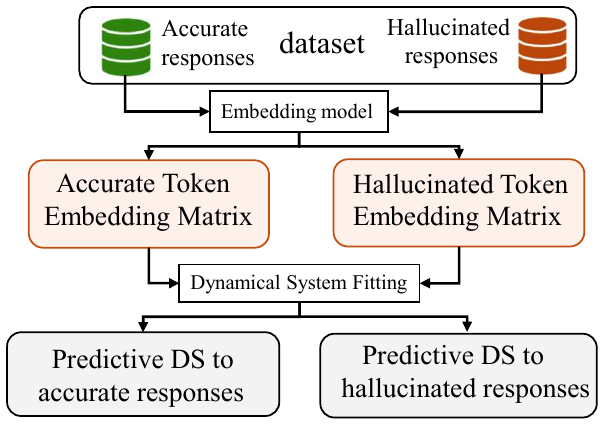}
    \label{fig:DS-train}}}
    \hfill
    \vspace{0.7cm}
    \subfloat[Phase 2: Hallucination detection]{{
    \includegraphics[width=0.44\textwidth]{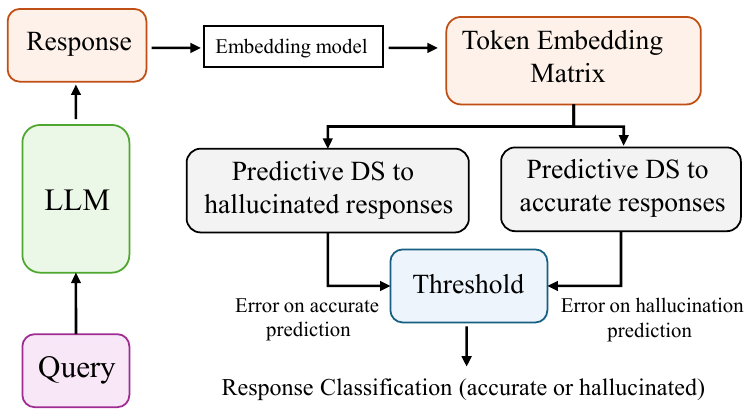}
    \label{fig:DS-inference}}}
    \vspace{-0.4cm}
    \caption{Hallucination Detection Dynamical System (DS): (a) Phase 1: Dynamical system fitting by embedding accurate and hallucinated datasets into token embedding matrices. This data is used to provide two different estimates for the action of the Koopman operator that describes the temporal evolution of the token embedding for accurate and hallucinated responses, (b) Phase 2: Hallucination detection of LLM responses to queries by embedding them and passing them through both predictive models. The classification is determined by comparing the relative accuracy of these estimates.}
    \label{fig:approach}
\end{figure*}

Our method illustrated in Fig.~\ref{fig:approach} proceeds in two phases. The offline DS fitting stage shown in Fig.~\ref{fig:DS-train} learns $i)$ a DS on non-hallucinated token embedding trajectories, and $ii)$ a DS on hallucinated token embedding trajectories. The inference stage depicted in Fig.~\ref{fig:DS-inference} classifies a previously unseen LLM response by considering the relative accuracy between predictions of the token embedding trajectories from models inferred from hallucinated and non-hallucinated responses.\vspace{-0.2cm}

\hspace{-0.4cm}
\subsection{Data-Driven Dynamical System Inference (Phase 1)}
Dynamic mode decomposition (DMD) is a well-established strategy for estimating a finite-dimensional approximation for the action of the Koopman operator from data \cite{kutz16, schm10, rowl09, will15, wilson23koopman}. Here we use the Extended DMD approach \cite{will15} which first `lifts' the observables to a higher dimensional space:
\begin{equation} \label{liftobs}
    \mathbf{z}_k = \begin{bmatrix} \mathbf{y}_k^\top & f_{\rm lift}^\top( \mathbf{y}_k)\end{bmatrix}^\top,
\end{equation}
where $\mathbf{z}_k \in \mathbb{R}^{M+\gamma}$ is the lifted observable with $f_{\rm lift} \in \mathbb{R}^\gamma$ where $\gamma$ being the dimension of the lifting. Above, $^\top$ indicates the vector transpose. Common choices of lifted coordinates include polynomial combinations of observables and radial basis functions \cite{will15}. This lifting yields a higher-dimensional set of observables which ultimately improves the accuracy of the subsequent model fitting. A series of snapshot pairs $s_k = (\mathbf{z}_k,\mathbf{z}_{k+1})$ is collected for $k = 1,\dots,q$ and the pairs are arranged into matrices $\bm{X} = \begin{bmatrix} \mathbf{z}_1 & \dots & \mathbf{z}_{q} \end{bmatrix}$ and $\bm{X}^+ = \begin{bmatrix} \mathbf{z}_2 & \dots & \mathbf{z}_{q+1} \end{bmatrix}$. A finite-dimensional approximation of the Koopman operator that captures the evolution of $\mathbf{z}_{k+1} = \mathbf{A} \mathbf{z}_k$ can be obtained taking
\begin{equation} \label{dmdfitting}
    \mathbf{A} = \mathbf{X}^+ \mathbf{X}^\dagger,
\end{equation}
where $^\dagger$ denotes the pseudoinverse. To avoid overfitting, low-rank fits to the data can also be obtained by truncating the SVD modes of $\mathbf{X}$ prior to taking the pseudoinverse \cite{proc16}.

From a dynamics systems perspective, we posit that correct and hallucinated responses from the LLM represent solutions evolving on different manifolds $\mathcal{M}_h$, $\mathcal{M}_c \in \mathbb{R}^N$, respectively. As such, data-driven Koopman-based model identification strategies applied to hallucinated and correct snapshot pairs will yield different estimates for the action of the Koopman operator \cite{budi12}. Snapshot pairs of lifted observables are placed in the matrices $\bm{X}_c$ and $\bm{X}^+_c$ or $\bm{X}_h$ and $\bm{X}^+_h$ depending on whether the LLM passages are hallucinations or not (as gauged by humans with separate annotations provided in the datasets). This data is used to approximate the action of the Koopman operator according to \eqref{dmdfitting} yielding two approximations: $\mathbf{z}_{k+1} = \mathbf{A}_c \mathbf{z}_k$ and $\mathbf{z}_{k+1} = \mathbf{A}_h \mathbf{z}_k$.

\subsection{Classification via Differential Residual Score (Phase 2)}
For an LLM response containing $L$ tokens, we obtain a set of lifted observables $\bm{z}_1, \dots, \bm{z}_L$ defined according to \eqref{liftobs}. We consider the error associated with the prediction of the evolution of the token embeddings from $\bm{y}_{k}$ to $\bm{y}_{k+1}$ according to
\begin{align}
    \bm{\epsilon}_{c,k} &\equiv || \bm{y}_{k+1} - \begin{bmatrix} \bm{I} & \bm{0} \end{bmatrix} \bm{A}_c \bm{z}_k|| , \nonumber \\
    \bm{\epsilon}_{h,k} &\equiv ||\bm{y}_{k+1} - \begin{bmatrix} \bm{I} & \bm{0} \end{bmatrix} \bm{A}_h \bm{z}_k||,
\end{align}
where $\bm{I} \in \mathbb{R}^{M \times M}$ denotes the identity matrix, $\bm{0} \in \mathbb{R}^{M \times \gamma}$ is a matrix of zeros, and $||\cdot||$ denotes the 2-norm. In the above definition, recall that the first $M$ rows of $\bm{z}_k$ comprise the observable $\bm{y}_k$ so that $\bm{\epsilon}_{c,k}$ and $\bm{\epsilon}_{h,k}$ do not consider the error associated with the prediction of the lifted coordinates. The prediction errors can subsequently be compared to determine whether $\bm{A}_c$ or $\bm{A}_h$ (capturing the behavior on $\mathcal{M}_c$ and $\mathcal{M}_h$, respectively) provide a better estimate of the dynamics. We then define
\begin{equation} \label{tokenlevel}
    \Delta \mathcal{E}_k = \bm{\epsilon}_{h,k} - \bm{\epsilon}_{c,k},
\end{equation}
as the token-level residual score and 
\begin{equation} \label{responselevel}
    \Delta \mathcal{E} = \bigg( \sum_{j=1}^{L-1} \bm{\epsilon}_{h,j}^2 \bigg)^{1/2} - \bigg( \sum_{j=1}^{L-1} \bm{\epsilon}_{c,j}^2 \bigg)^{1/2},
\end{equation}
as the response-level residual score. Intuitively, when the output of an LLM is (resp.,~is not) a hallucination, $\bm{A}_h$ (resp.,~$\bm{A}_c$) should yield a better prediction biasing $\Delta \mathcal{E}_k$ and $\Delta \mathcal{E}$ towards negative (resp.,~positive) values. To yield a binary classification $\widehat{D} \in \{0, 1\}$, where $1$ denotes a hallucination and $0$ denotes factual text, we apply a decision threshold $\eta$:
\begin{equation} \label{decisioneq}
    \widehat{D} = 
    \begin{cases} 
    1, & \text{if } \Delta \mathcal{E} < \eta, \\ 
    0, & \text{if } \Delta \mathcal{E} \geq \eta. 
    \end{cases}
\end{equation}
The threshold $\eta$ serves as a hyperparameter to tune the balance between precision and recall, allowing for the optimization of the $F_1$ score across various LLM architectures. The length of the response in \eqref{responselevel} can be adjusted as desired to classify individual sentences produced by the LLM or entire passages in response to a user prompt.
\hspace{-0.2cm}
\section{Experiments and Results}
\subsection{Experimental Setup}
\textbf{Datasets.} To evaluate the performance of our DS approach, we benchmark on three diverse datasets that represent varying levels of hallucination granularity:
\begin{itemize}[leftmargin=*]
\item \texttt{WikiBio} \cite{manakul2023selfcheckgpt} contains 238 LLM-generated biographies, totaling 1,908 sentences, each manually annotated for factual drift. Sentences are categorized into three classes: major inaccurate, minor inaccurate, and accurate.
\item The summarization task of the \texttt{HaluEval} dataset \cite{li2023halueval} provides 10K pairs of correct and hallucinated responses annotated at the summary level only.
\item \texttt{FELM} \cite{zhao2023felm} focuses on sentence-level reasoning across specialized domains (Math, Science, and Logic). It has 4427 manually labeled sentences with 3642 being correct and 785 being hallucinated.
\end{itemize}

\textbf{Embedding Models.} We test our DS method using five top-performing embedding models from the HuggingFace leaderboard of the \href{https://huggingface.co/spaces/mteb/leaderboard}{Massive Text Embedding Benchmark} (MTEB) as described in Table~\ref{tab:model_summary}. Working within limited computing resources, we selected models of varying sizes, from the lightweight $30$M-parameter \texttt{Jina-v5-Nano} to the larger $8$B-parameter \texttt{Llama-3}, thereby balancing performance with practical hardware constraints.
\begin{table}[h!]
\caption{Benchmarked embedding models for DS hallucination detection.}
\label{tab:model_summary}
\centering
\resizebox{\columnwidth}{!}{
\begin{tabular}{@{}lllll@{}}
\toprule
\textbf{Model} & \textbf{Parameters} & \textbf{Dim} ($\bm{M}$)& \textbf{Max Context} & \textbf{Architecture Type} \\ \midrule
\texttt{Jina-v5-Nano} \cite{akram2026jina} & $0.03$\,B & $768$ & $8192$ & Bi-directional Encoder \\
\texttt{Qwen3-Embed} \cite{li2026qwen3} & $0.6$\,B & $1024$ & $32768$ & Dense Decoder-Only \\
\texttt{F2LLM} \cite{zhang2026f2llm}& $4.0$\,B & $2560$ & $4096$ & Hybrid (Code/NL) \\
\texttt{Mistral} \cite{jiang2023mistral7b}& $7.2$\,B & $4096$ & $32768$ & Sparse Attention (SMoE) \\
\texttt{Llama-3} \cite{grattafiori2024llama}& $8.0$\,B & $4096$ & $8192$ & Dense Causal Decoder \\
\bottomrule
\end{tabular}
}
\end{table}

\subsection{Implementation Details}
\subsubsection{Embedding Extraction}
Embedding extraction for all datasets relies on capturing the embedding per token for a given sentence. For encoder-based models like \texttt{Jina-v5} and \texttt{Qwen3-Embed}, we extract the last hidden state of the final layer which represents each token as a high-dimensional vector. For decoder-only models such as \texttt{F2LLM} and \texttt{Mistral}, we explicitly enable the parameter \texttt{output\_hidden\_states=True} and get the activation trajectory from the final transformer block. In the case of \texttt{Llama-3}, we utilize the \texttt{LLM2Vec} wrapper which patches the causal attention mask to allow for bidirectional context and ensures that each embedding in the sequence is informed by both preceding and succeeding tokens. Across all models, the resulting tensors are of size \texttt{embedding dimension} $\times$ \texttt{number of tokens} format.
\subsubsection{Dynamical System Fitting}
Because the dimension of each embedding varies between models, we take the observable from \eqref{dynsyst} to be $\bm{y}_k = \bm{\Phi}^\top H(\mathbf{x}_k)$ where $\mathbf{\Phi} \in \mathbb{R}^{M \times 500}$ is comprised of the 500 most dominant SVD modes from the fitting data, as gauged by the singular values of the covariance matrix of the fitting data. This standardizes the dimension of the observables and allows for the investigation of cross-embedding generalization (Section~\ref{crosssec}). Taking the observable to be $H(\mathbf{x}_k)$ results in negligible differences in the classification accuracy. The lifting function $f_{\rm lift}$ takes polynomial combinations up to order 4 of a subset of the most dominant SVD modes. For the \texttt{HaluEval} dataset, 2000 entries are used for fitting the data-driven models with the remaining 8000 datasets used for testing. The \texttt{WikiBio} and \texttt{FELM} datasets have substantially fewer samples necessitating the use of a larger portion for fitting. For the \texttt{WikiBio} dataset, 206 entries are used for fitting and 32 are used for testing. For the \texttt{FELM} dataset, 746 entries are used for fitting and 100 total datasets (with 20 each from the science, world knowledge, reasoning, math, and writing categories) are used for testing. For each dataset, the division of the fitting and testing samples does not meaningfully alter the results. AUC curves are obtained by sweeping $\eta$ from \eqref{decisioneq} over plausible values and computing true/false positives and negatives over this sweep. $F_1$ score and balanced accuracy is reported for $\eta$ that maximizes the reported metric.

\subsection{Main Results}

\subsubsection{Dynamical System Classification Improves Hallucination Detection}
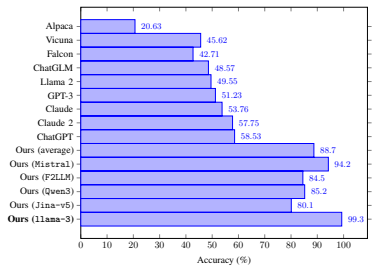
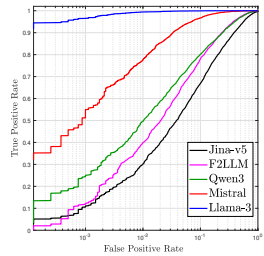
\begin{figure}[!b]
    \centering
    \vspace{-0.6cm}
    \hspace{-1.2cm}
    \subfloat[\footnotesize{Accuracy against other LLMs} \label{fig:acc-halu}]{
    \begin{tikzpicture}[scale=0.6]
    \begin{axis}[
        xbar,
        xlabel={Accuracy (\%)},
        ytick={0,...,14},
        yticklabels={\textbf{Ours (\texttt{\textbf{llama-3}})}, Ours (\texttt{Jina-v5}), Ours (\texttt{Qwen3}), Ours (\texttt{F2LLM}), Ours (\texttt{Mistral}), Ours (average), ChatGPT, Claude 2, Claude, GPT-3, Llama 2, ChatGLM, Falcon, Vicuna, Alpaca},
        xmin=0, 
        xmax=112,
        enlarge x limits={value=0.01, upper},
        xlabel style={align=center},
        yticklabel style={text width=3.2cm, align=right, font=\small},
        bar width=0.45cm,
        width=11cm,
        height=9cm,
        nodes near coords,
        nodes near coords align={horizontal},
        every node near coord/.append style={anchor=west, xshift=3pt, font=\footnotesize} 
    ]
    \addplot coordinates {
        (99.3,0) (80.1,1) (85.2,2) (84.5,3) (94.2,4)
        (88.7,5) (58.53,6) (57.75,7) (53.76,8) (51.23,9)
        (49.55,10) (48.57,11) (42.71,12) (45.62,13) (20.63,14)
    };
    \end{axis}
    \end{tikzpicture}
    }
    \hfil
    \hspace{-0.3cm}
    \subfloat[\footnotesize{ROC curves for all embedding models}]{{
    \includegraphics[width=0.4\textwidth]{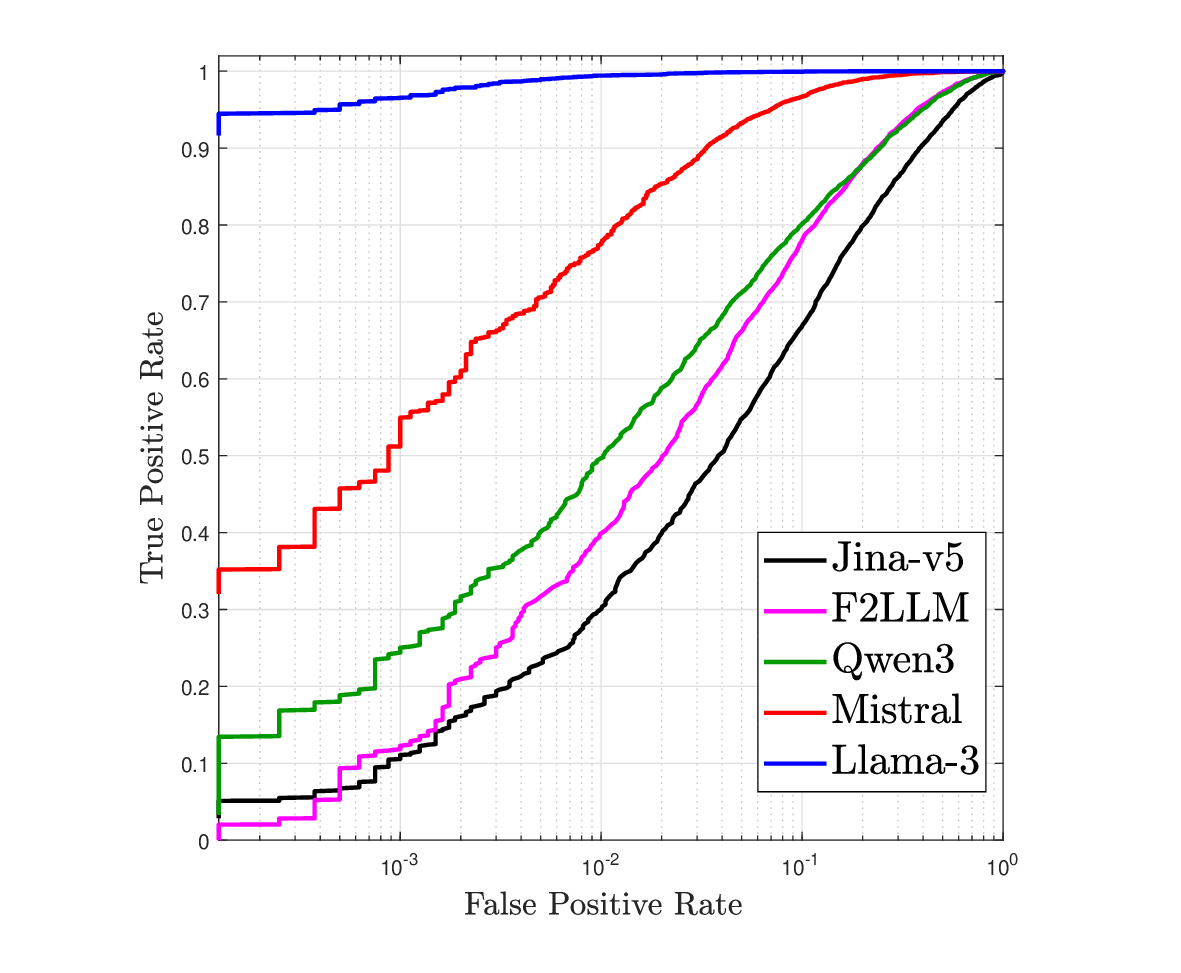}
    \label{fig:roc-halu}}}    
    \caption{Hallucination detection performance using DS classification on 8K samples of the HaluEval summarization task, showing (a) classification accuracy for multiple black-box LLMs and (b) the corresponding ROC curves for all embedding models.}
    \vspace{-0.2cm}
    \label{fig:classif-Halu}
\end{figure}

We begin by analyzing the performance of our method on the summarization task of the \texttt{HaluEval} dataset. Fig.~\ref{fig:acc-halu} illustrates the comparative accuracy across models, showing how our DS approach achieves high classification accuracy across a diverse range of embedding scales, even for the highly compressed \texttt{0.03B Jina-v5} model. This confirms our hypothesis on how the factual drift is fundamentally encoded within the embedding transitions across tokens. The ROC curves in Fig.~\ref{fig:roc-halu} further validate this discriminative capability of our method across embedding models. The \texttt{Llama-3} and \texttt{Mistral} models exhibit the highest area under the curve (AUC), indicating that increased parameter count and embedding dimensionality provide a clear separation of generative regimes. The performance of the smaller models highlights the efficiency of our differential residual score $\Delta\mathcal{E}$ as they are already outperforming zero-shot hallucination LLM detectors with a single-sample pass and without the need for multiple stochastic samples or external verification.

We then investigate whether our DS-based hallucination detector can identify the factuality of sentences on the \texttt{WikiBio} dataset. In detecting non-factual sentences, both \texttt{major-inaccurate} labels and \texttt{minor-inaccurate} labels are grouped together into the \textit{non-factual} class, while the \textit{factual} class refers to accurate sentences. In addition, we consider the task of detecting major-inaccurate sentences in passages that are \textit{not} total hallucination passages, which we refer to as \textit{non-factual}$^*$. Table~\ref{tab:selfcheckgpt-aucpr} reports hallucination detection performance measured in AUC-PR since the data is unbalanced. For the primary NonFact task, all of our methods outperform both the random baseline and \texttt{SelfCheckGPT}, with \texttt{llama-3} achieving the best result. The more challenging NonFact$^*$ task proves substantially more difficult across all methods, as reflected by the lower scores. Here, \texttt{Mistral} delivers the strongest performance and significantly surpasses \texttt{SelfCheckGPT} \cite{manakul2023selfcheckgpt}. On the Factual class, performance is generally lower than NonFact across all methods, with \texttt{SelfCheckGPT} slightly ahead of our variants.

Table~\ref{tab:felm-acc-halu} compares hallucination detection performance on the \texttt{FELM} dataset, reporting $F_1$ score and balanced accuracy for three baseline LLMs alongside our five embedding-based methods. \texttt{GPT-4} achieves the highest $F_1$ score. On balanced accuracy, our DS method using \texttt{Llama-3} embedding achieves the top score, while \texttt{Qwen3} also matches \texttt{GPT-4}.

\begin{table*}[!t]
    \centering
    \begin{minipage}{0.48\textwidth}
        \centering
        \caption{Hallucination detection performance on the \texttt{WikiBio} dataset.}
        \label{tab:selfcheckgpt-aucpr}
        \vskip 3pt
        \small
        \begin{tabular}{lccc}
            \toprule
            \multirow{2}{*}{Method} & \multicolumn{3}{c}{Sentence-level (AUC-PR)} \\
            \cmidrule(lr){2-4}
                                    & NonFact & NonFact* & Factual \\
            \midrule
            Random                  & 73.0   & 29.7     & 27.0 \\
            \midrule
            SelfCheckGPT            & 81.96  & 45.96    & {\bf 44.23} \\
            \midrule
            Ours (\texttt{llama-3}) & {\bf85.3} & 51.2 & 42.4 \\
            Ours (\texttt{Jina-v5}) & 82.2   & 44.5 & 37.2 \\
            Ours (\texttt{Qwen3})   & 84.7 & 47.5 & 37.6 \\
            Ours (\texttt{F2LLM})   & 82.2   & 44.3 & 42.0 \\
            Ours (\texttt{Mistral}) & 82.8 & {\bf 54.1} & 38.1 \\
            Ours (avg)              & 83.4      & 48.3       & 39.5 \\
            \bottomrule
        \end{tabular}
    \end{minipage}
    \hfill
    \begin{minipage}{0.48\textwidth}
        \vspace{-0.3cm}
        \centering
        \caption{Comparison of hallucination detection methods on the \texttt{FELM} dataset.}
        \label{tab:felm-acc-halu}
        \vskip 3pt
        \small
        \begin{tabular}{l c c}
            \toprule
            & $F_1$ Score & Bal. Acc. \\
            \midrule
            \texttt{gpt4-0314}      & {\bf 48.3} & 67.1 \\
            \texttt{Vicuna-33B}     & 32.5 & 56.5  \\
            \texttt{gpt3.5-0301}    & 25.5 & 55.9  \\
            \midrule
            Ours (\texttt{Jina-v5})  & 39.3 & 63.5  \\
            Ours (\texttt{F2LLM})   & 41.4 & 65.4 \\
            Ours (\texttt{Qwen3})   & 44.8 & 67.7 \\
            Ours (\texttt{Mistral}) & 44.7 & 67.0 \\
            Ours (\texttt{Llama-3}) & 43.0 & {\bf 68.4} \\
            \bottomrule
        \end{tabular}
    \end{minipage}
\end{table*}

\subsubsection{Detection Performance Improves with Longer Token Sequences}
\begin{figure*}[!t]
\vspace{-0.4cm}
\centering
    \subfloat[Average over models]{{
    \includegraphics[width=0.3\textwidth]{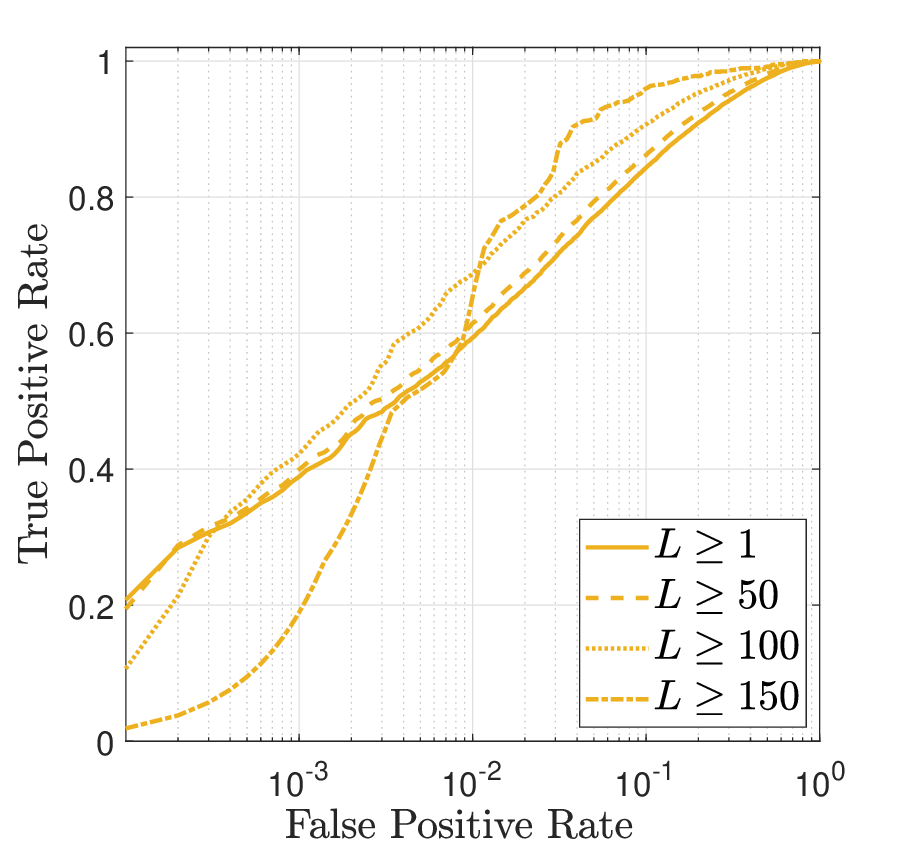}
    \label{fig:roc-average}}}
    \hfill
    \subfloat[\texttt{Jina-v5-Nano}]{{
    \includegraphics[width=0.3\textwidth]{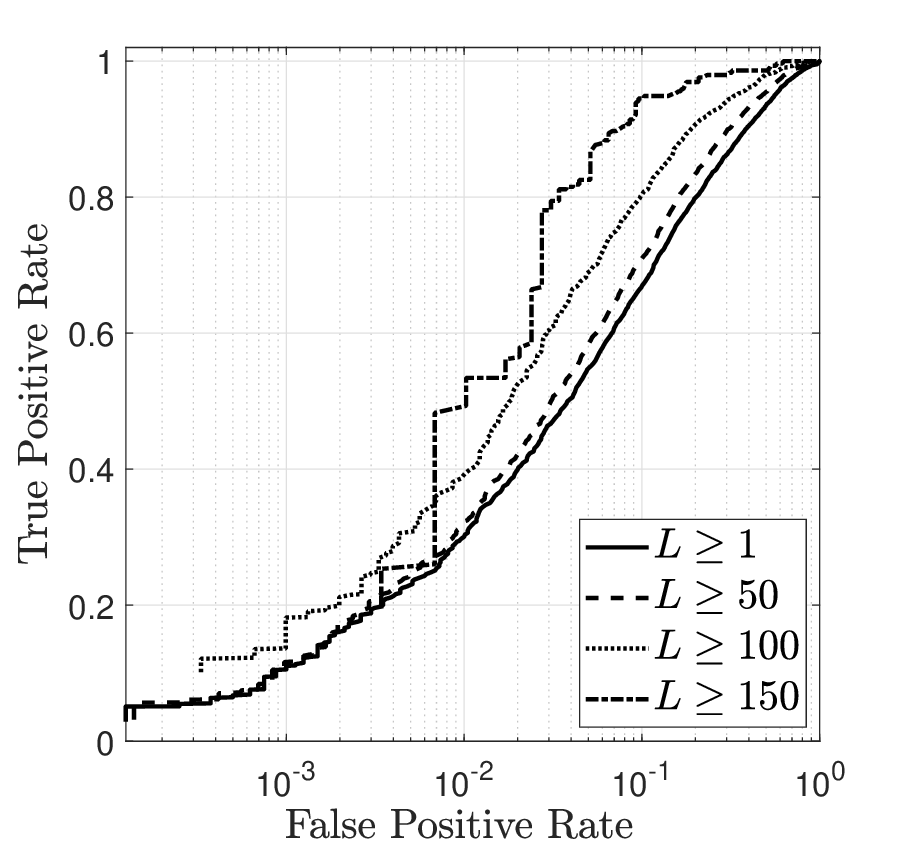}
    \label{fig:roc-jina}}}
    \hfill
    \subfloat[\texttt{Qwen3}]{{
    \includegraphics[width=0.3\textwidth]{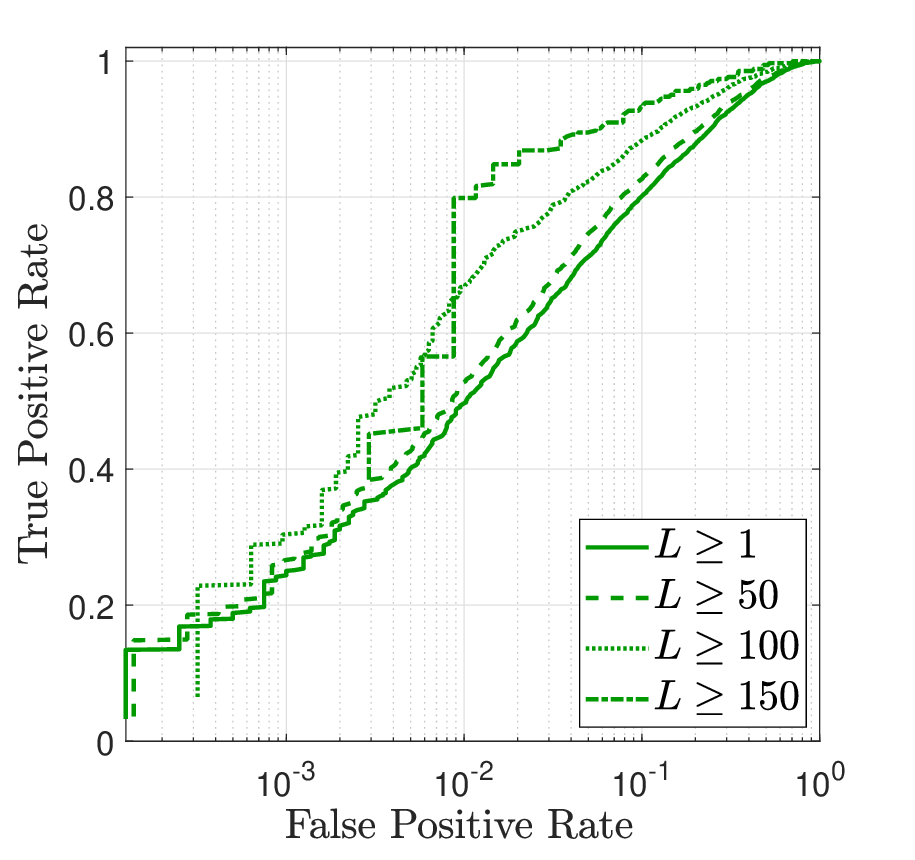}
    \label{fig:roc-qwen}}}
    \\
    \subfloat[\texttt{F2LLM}]{{
    \includegraphics[width=0.3\textwidth]{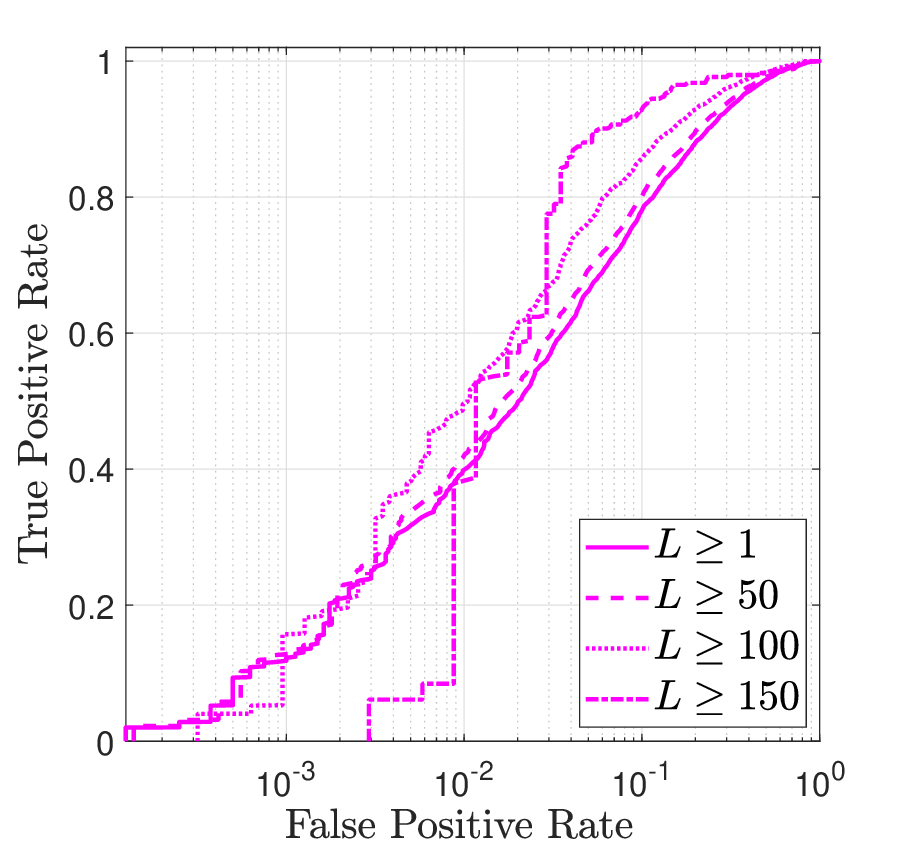}
    \label{fig:roc-f2llm}}}
    \hfill
    \subfloat[\texttt{Mistral}]{{
    \includegraphics[width=0.3\textwidth]{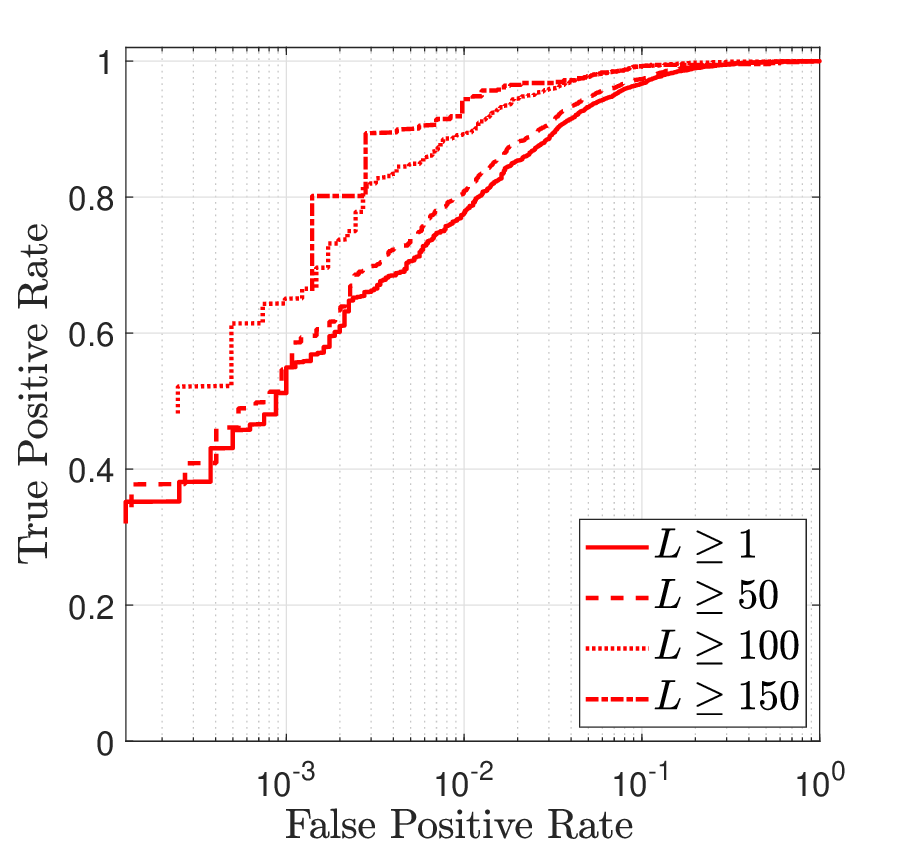}
    \label{fig:roc-mistral}}}
    \hfill
    \subfloat[\texttt{Llama-3}]{{
    \includegraphics[width=0.3\textwidth]{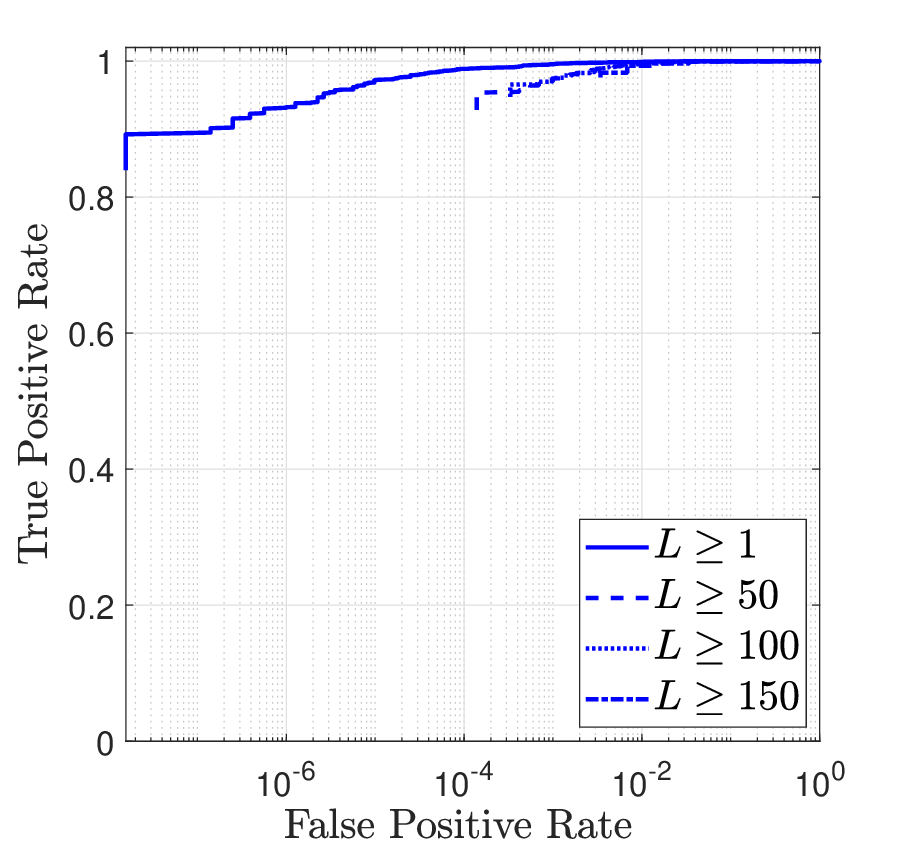}
    \label{fig:roc-llama3}}}
    \caption{ROC curves on the \texttt{HaluEval} dataset as a function of the sequence length $L$ for (a) the average classification performance and for (b)--(f) each embedding model showing the variation of the true positive rate against the false positive rate at various prediction error thresholds.}
    \label{fig:Roc-HaluEval-models-seqlength}
\end{figure*}

A central hypothesis of our DS approach is that longer token sequences yield more reliable hallucination detection, as they provide richer semantic trajectories. Indeed, for a semantic trajectory of length $L$, it has been shown that the prediction residual $\epsilon_k = \|\mathbf{x}_{k+1} - \mathbf{A}\mathbf{x}_k\|$ converges at a rate of $O(1/\sqrt{L})$ under a stable semantic manifold \cite{zhang2023quantitative}. To validate this hypothesis, we evaluate our DS hallucination detection method on the summarization samples of the \texttt{HaluEval} dataset, by varying the minimum sequence length threshold $L \in \{1, 50, 100, 150\}$. The ROC curves in Fig.~\ref{fig:Roc-HaluEval-models-seqlength} illustrate the effect of increasing the minimum sequence length $L$ on detection performance.

Notably, the magnitude of the improvement from shorter to longer sequence length correlates inversely with model size. Smaller embedding models, such as \texttt{Jina-v5} (Fig.~\ref{fig:roc-jina}), exhibit the most substantial gains, with the ROC curve transitioning from a gradual ascent at $L \geq 1$ to a sharp rise that hugs the upper boundary of the plot at $L \geq 150$. A similar pattern is observed for \texttt{Qwen3} (Fig.~\ref{fig:roc-qwen}) and \texttt{F2LLM} (Fig.~\ref{fig:roc-f2llm}), where the separation between the curves for $L \geq 1$ and $L \geq 150$ is clearly visible across the entire range of false positive rates. The average over models (Fig.~\ref{fig:roc-average}) summarizes this behavior, showing a consistent leftward and upward shift with increasing $L$. For larger models such as Mistral (Fig.~\ref{fig:roc-mistral}) and Llama-3 (Fig.~\ref{fig:roc-llama3}), the improvements are more subtle, as their embedding spaces already capture sufficiently discriminative semantic trajectories at shorter sequence lengths.

\begin{table*}[t!]
    \centering
    \caption{Cross-embedding generalization across multiple datasets. Rows represent the embedding model used to fit the DS parameters (Fit), and columns represent the target embedding model for evaluation (Test). Each cell reports balanced accuracy.}
    \label{tab:generalization_performance}
    \resizebox{\textwidth}{!}{%
    \begin{tabular}{l c c c c c c c c c c c c c c c}
        \toprule
        & \multicolumn{5}{c}{\texttt{HaluEval}} & \multicolumn{5}{c}{\texttt{FELM}} & \multicolumn{5}{c}{\texttt{WikiBio}} \\
        \cmidrule(lr){2-6} \cmidrule(lr){7-11} \cmidrule(lr){12-16}
        \textbf{Fit $\downarrow$ \ Test $\rightarrow$}
        & \texttt{Jina} & \texttt{F2} & \texttt{Qw3} & \texttt{Mis} & \texttt{Ll3}
        & \texttt{Jina} & \texttt{F2} & \texttt{Qw3} & \texttt{Mis} & \texttt{Ll3}
        & \texttt{Jina} & \texttt{F2} & \texttt{Qw3} & \texttt{Mis} & \texttt{Ll3} \\
        \midrule
        \texttt{Jina-v5} & 80.1 & \cellcolor{gray!20} 54.6 & \cellcolor{gray!20} 56.7 & \cellcolor{gray!20} 49.4 & \cellcolor{gray!20} 51.3 & 63.5 & \cellcolor{gray!20} 52.7 & \cellcolor{gray!20} 50.4 & \cellcolor{gray!20} 52.8 & \cellcolor{gray!20} 53.5 & 65.2 & \cellcolor{gray!20} 50.3 & \cellcolor{gray!20} 56.1 & \cellcolor{gray!20} 52.1 & \cellcolor{gray!20} 53.9 \\
        \texttt{F2LLM}   & \cellcolor{gray!20} 54.4 & 84.6 & \cellcolor{gray!20} 50.5 & \cellcolor{gray!20} 50.7 & \cellcolor{gray!20} 57.6 & \cellcolor{gray!20} 53.0 & 65.4 & \cellcolor{gray!20} \textbf{56.8} & \cellcolor{gray!20} 50.0 & \cellcolor{gray!20} 53.9 & \cellcolor{gray!20} 51.7 & 65.7 & \cellcolor{gray!20} 52.0 & \cellcolor{gray!20} 50.0 & \cellcolor{gray!20} 54.9 \\
        \texttt{Qwen3}   & \cellcolor{gray!20} 51.9 & \cellcolor{gray!20} \textbf{63.0} & 85.2 & \cellcolor{gray!20} 51.4 & \cellcolor{gray!20} 60.4 & \cellcolor{gray!20} 50.6 & \cellcolor{gray!20} 50.2 & 67.7 & \cellcolor{gray!20} 54.0 & \cellcolor{gray!20} 54.8 & \cellcolor{gray!20} 54.6 & \cellcolor{gray!20} 50.6 & 66.4 & \cellcolor{gray!20} 50.0 & \cellcolor{gray!20} 56.8 \\
        \texttt{Mistral} & \cellcolor{gray!20} 53.9 & \cellcolor{gray!20} 54.8 & \cellcolor{gray!20} 55.4 & 94.2 & \cellcolor{gray!20} 53.4 & \cellcolor{gray!20} 52.2 & \cellcolor{gray!20} 54.9 & \cellcolor{gray!20} 51.8 & 67.0 & \cellcolor{gray!20} 52.6 & \cellcolor{gray!20} \textbf{59.7} & \cellcolor{gray!20} 52.2 & \cellcolor{gray!20} 59.0 & 69.3 & \cellcolor{gray!20} 55.1 \\
        \texttt{Llama-3} & \cellcolor{gray!20} 51.8 & \cellcolor{gray!20} 51.3 & \cellcolor{gray!20} 50.9 & \cellcolor{gray!20} 52.0 & 99.3 & \cellcolor{gray!20} 50.5 & \cellcolor{gray!20} 50.0 & \cellcolor{gray!20} 50.0 & \cellcolor{gray!20} 50.0 & 68.3 & \cellcolor{gray!20} 52.5 & \cellcolor{gray!20} 56.5 & \cellcolor{gray!20} 57.0 & \cellcolor{gray!20} 52.9 & 65.6 \\
        \bottomrule
    \end{tabular}
    }
\end{table*}

Table~\ref{tab:sequence-length-analysis} corroborates these observations. Across all embedding models, the $F_1$ score, balanced accuracy, and AUC improve consistently as the minimum sequence length increases, with the most substantial gains observed for smaller models. \texttt{Llama-3} maintains near-perfect performance across all sequence lengths, confirming that its richer embedding space compensates for shorter sequences. This reveals a trade-off between model size and sequence length, suggesting that the representational limitations of smaller embedding models can be mitigated by increasing the observed token context.

\begin{table}[!t]
    \centering
    \caption{Performance of our DS hallucination detection method on 8K summarization samples of the \texttt{HaluEval} dataset over the number of tokens $L$. The value of each cell denotes $F_1$ score/balanced accuracy/AUC.}
    \label{tab:sequence-length-analysis}
    \resizebox{\columnwidth}{!}{
    \begin{tabular}{l c c c c}
        \toprule
        & \textbf{$L \geq 1$} & \textbf{$L \geq 50$} & \textbf{$L \geq 100$} & \textbf{$L\geq 150$} \\
        \midrule
        \texttt{Jina-v5} & 80.2\;/\;80.1\;/\;87.7  & 82.1\;/\;82.0\;/\;89.8  & 86.3\;/\;86.0\;/\;93.2  & 92.6\;/\;92.5\;/\;96.6  \\
        \texttt{F2LLM}   & 84.6\;/\;84.5\;/\;92.2  & 85.5\;/\;85.5\;/\;93.0  & 88.1\;/\;88.0\;/\;94.7  & 92.0\;/\;92.1\;/\;96.0  \\
        \texttt{Qwen3}   & 85.0\;/\;85.2\;/\;92.7  & 86.3\;/\;86.4\;/\;93.6  & 89.1\;/\;89.3\;/\;95.7  & 92.3\;/\;92.6\;/\;97.4  \\
        \texttt{Mistral} & 94.2\;/\;94.2\;/\;98.6  & 94.9\;/\;94.9\;/\;98.9  & 96.5\;/\;96.5\;/\;99.5  & 97.3\;/\;97.3\;/\;99.4  \\
        \texttt{Llama-3} & \textbf{99.3}\;/\;\textbf{99.3}\;/\;\textbf{99.9}  & \textbf{99.4}\;/\;\textbf{99.4}\;/\;\textbf{99.9}  & \textbf{99.4}\;/\;\textbf{99.4}\;/\;\textbf{99.9} & \textbf{99.3}\;/\;\textbf{99.3}\;/\;\textbf{99.9} \\
        \bottomrule
    \end{tabular}
    }
    \vspace{-0.2cm}
\end{table}

\subsubsection{Cross-Embedding Generalization of Dynamical Predictions} \label{crosssec}
Table~\ref{tab:generalization_performance} evaluates the cross-embedding generalization of our method across the three datasets. Each row corresponds to the embedding model used to fit the DS parameters and each column corresponds to the target embedding model on which the fitted system is evaluated, with each cell reporting balanced accuracy. The diagonal entries represent the matched setting, where the same embedding model is used for both fitting and evaluation. From a dynamical systems perspective, transfer between models is surprising. The approximations $\bm{A}_h$ and $\bm{A}_c$ are inferred for a specific embedding using the extended DMD algorithm \cite{will15}. When applying this method, in general, even permutations to the ordering of the observables will completely ruin the predictive ability of the resulting model requiring the matrices to be refit to the data. There is no obvious reason that a model fit using one set of observables could give information about the dynamics of a model with completely different observables. Nonetheless, many off-diagonal entries in Table~\ref{tab:generalization_performance} exceed the $50\%$ chance level by a non-trivial margin. For instance, on \texttt{HaluEval}, fitting on \texttt{Qwen3} and testing on \texttt{F2LLM} yields $63.0$\%, and fitting on \texttt{Qwen3} and testing on \texttt{Llama-3} reaches $60.4\%$. Similar patterns appear on \texttt{WikiBio}, where fitting on \texttt{Mistral} and testing on \texttt{Jina-v5} achieves $59.7\%$, and fitting on \texttt{Mistral} and testing on \texttt{Qwen3} reaches $59.0\%$. Recalling that the observables used for model identification are the magnitudes of the most dominant SVD modes of the token embeddings, there may be differences in power law scalings present between hallucinated and correct responses. This partial transferability suggests that, despite operating in distinct embedding spaces, certain embedding models share overlapping manifold properties in how they represent semantic evolution. This suggests that the dynamical signatures of embedding factuality and hallucination have universal properties of the observation manifold rather than model-specific artifacts.

\section{Conclusion}
In this paper, we introduce a new hallucination detection method by rethinking a Large Language Model as a black-box dynamical system where the temporal evolution of trajectories in the output space can reveal whether the text is correct or hallucinated. Our classification strategy thresholds a differential residual score $\Delta\mathcal{E}$ that corresponds to the prediction error between two dynamical systems fitted in the embedding space on correct or hallucinated samples. Empirical results demonstrate that our method consistently matches or outperforms state-of-the-art black-box baselines, thereby providing a low-cost alternative to existing detection methods.

\bibliographystyle{IEEEtran}
\bibliography{refs}

@inproceedings{manakul2023selfcheckgpt,
  title={Selfcheckgpt: Zero-resource black-box hallucination detection for generative large language models},
  author={Manakul, Potsawee and Liusie, Adian and Gales, Mark},
  booktitle={Proceedings of the 2023 conference on empirical methods in natural language processing},
  pages={9004--9017},
  year={2023}
}

@inproceedings{li2023halueval,
  title={Halueval: A large-scale hallucination evaluation benchmark for large language models},
  author={Li, Junyi and Cheng, Xiaoxue and Zhao, Wayne Xin and Nie, Jian-Yun and Wen, Ji-Rong},
  booktitle={Proceedings of the 2023 conference on empirical methods in natural language processing},
  pages={6449--6464},
  year={2023}
}

@article{zhao2023felm,
  title={Felm: Benchmarking factuality evaluation of large language models},
  author={Zhao, Yiran and Zhang, Jinghan and Chern, I and Gao, Siyang and Liu, Pengfei and He, Junxian and others},
  journal={Advances in Neural Information Processing Systems},
  volume={36},
  pages={44502--44523},
  year={2023}
}

@article{chen2024inside,
  title={{INSIDE}: {LLM}s' internal states retain the power of hallucination detection},
  author={Chen, Chao and Liu, Kai and Chen, Ze and Gu, Yi and Wu, Yue and Tao, Mingyuan and Fu, Zhihang and Ye, Jieping},
  journal={arXiv preprint arXiv:2402.03744},
  year={2024}
}

@article{zhang2023quantitative,
  title={A quantitative analysis of Koopman operator methods for system identification and predictions},
  author={Zhang, Christophe and Zuazua, Enrique},
  journal={Comptes Rendus. M{\'e}canique},
  volume={351},
  number={S1},
  pages={1--31},
  year={2023}
}

@article{kalai2025language,
  title={Why language models hallucinate},
  author={Kalai, Adam Tauman and Nachum, Ofir and Vempala, Santosh S and Zhang, Edwin},
  journal={arXiv preprint arXiv:2509.04664},
  year={2025}
}

@inproceedings{kalai2024calibrated,
  title={Calibrated language models must hallucinate},
  author={Kalai, Adam Tauman and Vempala, Santosh S},
  booktitle={Proceedings of the 56th Annual ACM Symposium on Theory of Computing},
  pages={160--171},
  year={2024}
}

@article{suzgun2025language,
  title={Language models cannot reliably distinguish belief from knowledge and fact},
  author={Suzgun, Mirac and Gur, Tayfun and Bianchi, Federico and Ho, Daniel E and Icard, Thomas and Jurafsky, Dan and Zou, James},
  journal={Nature Machine Intelligence},
  pages={1--11},
  year={2025},
  publisher={Nature Publishing Group UK London}
}

@article{huang2025survey,
  title={A survey on hallucination in large language models: Principles, taxonomy, challenges, and open questions},
  author={Huang, Lei and Yu, Weijiang and Ma, Weitao and Zhong, Weihong and Feng, Zhangyin and Wang, Haotian and Chen, Qianglong and Peng, Weihua and Feng, Xiaocheng and Qin, Bing and others},
  journal={ACM transactions on information systems},
  volume={43},
  number={2},
  pages={1--55},
  year={2025},
  publisher={ACM New York, NY}
}

@inproceedings{bender2021dangers,
  title={On the dangers of stochastic parrots: Can language models be too big?},
  author={Bender, Emily M and Gebru, Timnit and McMillan-Major, Angelina and Shmitchell, Shmargaret},
  booktitle={Proceedings of the 2021 ACM conference on fairness, accountability, and transparency},
  pages={610--623},
  year={2021}
}

@inproceedings{lin2022truthfulqa,
  title={Truthful{QA}: Measuring how models mimic human falsehoods},
  author={Lin, Stephanie and Hilton, Jacob and Evans, Owain},
  booktitle={Proceedings of the 60th annual meeting of the association for computational linguistics (volume 1: long papers)},
  pages={3214--3252},
  year={2022}
}

@article{li2023benchmarking,
  title={Benchmarking large language models for news summarization},
  author={Li, Junyi and Tang, Tianyi and Zhao, Wayne Xin and Wen, Ji-Rong},
  journal={arXiv preprint arXiv:2305.09034},
  year={2023}
}

@inproceedings{goel2025zero,
  title={Zero-knowledge {LLM} hallucination detection and mitigation through fine-grained cross-model consistency},
  author={Goel, Aman and Schwartz, Daniel and Qi, Yanjun},
  booktitle={Proceedings of the 2025 Conference on Empirical Methods in Natural Language Processing: Industry Track},
  pages={1982--1999},
  year={2025}
}

@article{brown2020language,
  title={Language models are few-shot learners},
  author={Brown, Tom and Mann, Benjamin and Ryder, Nick and Subbiah, Melanie and Kaplan, Jared D and Dhariwal, Prafulla and Neelakantan, Arvind and Shyam, Pranav and Sastry, Girish and Askell, Amanda and others},
  journal={Advances in neural information processing systems},
  volume={33},
  pages={1877--1901},
  year={2020}
}

@article{achiam2023gpt,
  title={{GPT}-4 technical report},
  author={Achiam, Josh and Adler, Steven and Agarwal, Sandhini and Ahmad, Lama and Akkaya, Ilge and Aleman, Florencia Leoni and Almeida, Diogo and Altenschmidt, Janko and Altman, Sam and Anadkat, Shyamal and others},
  journal={arXiv preprint arXiv:2303.08774},
  year={2023}
}

@article{warraich2025ethical,
  title={Ethical Governance of Artificial Intelligence Hallucinations in Legal Practice},
  author={Warraich, Muhammad Khurram Shahzad and Usman, Hazrat and Zakir, Sidra and Mehboob, Mohaddas},
  journal={Social Sciences Spectrum},
  volume={4},
  number={2},
  pages={603--615},
  year={2025}
}

@inproceedings{azaria2023internal,
  title={The internal state of an {LLM} knows when it’s lying},
  author={Azaria, Amos and Mitchell, Tom},
  booktitle={Findings of the Association for Computational Linguistics: EMNLP 2023},
  pages={967--976},
  year={2023}
}

@inproceedings{su2024unsupervised,
  title={Unsupervised real-time hallucination detection based on the internal states of large language models},
  author={Su, Weihang and Wang, Changyue and Ai, Qingyao and Hu, Yiran and Wu, Zhijing and Zhou, Yujia and Liu, Yiqun},
  booktitle={Findings of the Association for Computational Linguistics: ACL 2024},
  pages={14379--14391},
  year={2024}
}

@inproceedings{bar2025learning,
  title={Learning on {LLM} output signatures for gray-box behavior analysis},
  author={Bar-Shalom, Guy and Frasca, Fabrizio and Lim, Derek and Gelberg, Yoav and Ziser, Yftah and El-Yaniv, Ran and Chechik, Gal and Maron, Haggai},
  booktitle={ICML 2025 Workshop on Reliable and Responsible Foundation Models},
  year={2025}
}

@inproceedings{qian2025beyond,
  title={Beyond the next token: Towards prompt-robust zero-shot classification via efficient multi-token prediction},
  author={Qian, Junlang and Zhu, Zixiao and Zhou, Hanzhang and Feng, Zijian and Zhai, Zepeng and Mao, Kezhi},
  booktitle={Proceedings of the 2025 Conference of the Nations of the Americas Chapter of the Association for Computational Linguistics: Human Language Technologies (Volume 1: Long Papers)},
  pages={7093--7115},
  year={2025}
}

@article{qiao2026lowest,
  title={Lowest Span Confidence: A Zero-Shot Metric for Efficient and Black-Box Hallucination Detection in {LLM}s},
  author={Qiao, Yitong and Pan, Licheng and Mi, Yu and Liu, Lei and Shen, Yue and Sun, Fei and Chu, Zhixuan},
  journal={arXiv preprint arXiv:2601.19918},
  year={2026}
}

@article{farquhar2024detecting,
  title={Detecting hallucinations in large language models using semantic entropy},
  author={Farquhar, Sebastian and Kossen, Jannik and Kuhn, Lorenz and Gal, Yarin},
  journal={Nature},
  volume={630},
  number={8017},
  pages={625--630},
  year={2024},
  publisher={Nature Publishing Group UK London}
}

@inproceedings{hu2024embedding,
  title={Embedding and gradient say wrong: A white-box method for hallucination detection},
  author={Hu, Xiaomeng and Zhang, Yiming and Peng, Ru and Zhang, Haozhe and Wu, Chenwei and Chen, Gang and Zhao, Junbo},
  booktitle={Proceedings of the 2024 Conference on Empirical Methods in Natural Language Processing},
  pages={1950--1959},
  year={2024}
}

@article{sriramanan2024llm,
  title={{LLM}-check: Investigating detection of hallucinations in large language models},
  author={Sriramanan, Gaurang and Bharti, Siddhant and Sadasivan, Vinu Sankar and Saha, Shoumik and Kattakinda, Priyatham and Feizi, Soheil},
  journal={Advances in Neural Information Processing Systems},
  volume={37},
  pages={34188--34216},
  year={2024}
}

@inproceedings{sawczyn2026factselfcheck,
  title={Factselfcheck: Fact-level black-box hallucination detection for {LLM}s},
  author={Sawczyn, Albert and Binkowski, Jakub and Janiak, Denis and Gabrys, Bogdan and Kajdanowicz, Tomasz Jan},
  booktitle={Findings of the Association for Computational Linguistics: EACL 2026},
  pages={5603--5621},
  year={2026}
}

@inproceedings{zhang2023sac3,
  title={{SAC}3: reliable hallucination detection in black-box language models via semantic-aware cross-check consistency},
  author={Zhang, Jiaxin and Li, Zhuohang and Das, Kamalika and Malin, Bradley and Kumar, Sricharan},
  booktitle={Findings of the Association for Computational Linguistics: EMNLP 2023},
  pages={15445--15458},
  year={2023}
}

@article{kong2025multiperspective,
  title={Multi-perspective consistency checking for large language model hallucination detection: a black-box zero-resource approach},
  author={Kong, Linggang and Zhong, Xiaofeng and Chen, Jie and Fu, Haoran and Wang, Yongjie},
  journal={Frontiers of Information Technology \& Electronic Engineering},
  volume={26},
  number={11},
  pages={2298--2309},
  year={2025},
  publisher={ZUP}
}

@inproceedings{quevedo2024detecting,
  title={Detecting hallucinations in large language model generation: A token probability approach},
  author={Quevedo, Ernesto and Salazar, Jorge Yero and Koerner, Rachel and Rivas, Pablo and Cerny, Tomas},
  booktitle={World Congress in Computer Science, Computer Engineering \& Applied Computing},
  pages={154--173},
  year={2024},
  organization={Springer}
}

@inproceedings{DBLP:conf/iclr/KuhnGF23,
  author       = {Lorenz Kuhn and
                  Yarin Gal and
                  Sebastian Farquhar},
  title        = {Semantic Uncertainty: Linguistic Invariances for Uncertainty Estimation
                  in Natural Language Generation},
  booktitle    = {The Eleventh International Conference on Learning Representations,
                  {ICLR} 2023, Kigali, Rwanda, May 1-5, 2023},
  publisher    = {OpenReview.net},
  year         = {2023},
  url          = {https://openreview.net/forum?id=VD-AYtP0dve},
  bibsource    = {dblp computer science bibliography, https://dblp.org}
}

@article{mezi13,
  title={Analysis of fluid flows via spectral properties of the {K}oopman operator},
  author={I. Mezi{\'c}},
  journal={Annual Review of Fluid Mechanics},
  volume={45},
  pages={357--378},
  year={2013},
  publisher={Annual Reviews}
}

@article{budi12,
  title={Applied {K}oopmanism},
  author={M. Budi{\v{s}}i{\'c} and R. Mohr and I. Mezi{\'c}},
  journal={Chaos: An Interdisciplinary Journal of Nonlinear Science},
  volume={22},
  number={4},
  pages={047510},
  year={2012},
  publisher={AIP}
}

@article{schm10,
  title={Dynamic mode decomposition of numerical and experimental data},
  author={P. J. Schmid},
  journal={Journal of Fluid Mechanics},
  volume={656},
  pages={5--28},
  year={2010},
  publisher={Cambridge University Press}
}

@article{will15,
  title={A data--driven approximation of the {k}oopman operator: Extending dynamic mode decomposition},
  author={M. O. Williams  and I. G. Kevrekidis and C. W. Rowley},
  journal={Journal of Nonlinear Science},
  volume={25},
  number={6},
  pages={1307--1346},
  year={2015},
  publisher={Springer}
}

@article{rowl09,
  title={Spectral analysis of nonlinear flows},
  author={C. W. Rowley and I. Mezic and S. Bagheri and P. Schlatter and D. S. Henningson},
  journal={Journal of Fluid Mechanics},
  volume={641},
  number={1},
  pages={115--127},
  year={2009},
  publisher={Citeseer}
}

@book{kutz16,
  title={Dynamic mode decomposition: data-driven modeling of complex systems},
  author={J. N. Kutz and S. L. Brunton and B. W. Brunton and J. L. Proctor},
  year={2016},
  publisher={Society for Industrial and Applied Mathematics},
  address={Philadelphia, PA}
}

@article{proc16,
  title={Dynamic mode decomposition with control},
  author={J. L. Proctor  and S. L.  Brunton and J. N. Kutz},
  journal={SIAM Journal on Applied Dynamical Systems},
  volume={15},
  number={1},
  pages={142--161},
  year={2016},
  publisher={SIAM}
}

@article{akram2026jina,
  title={jina-embeddings-v5-text: Task-Targeted Embedding Distillation},
  author={Akram, Mohammad Kalim and Sturua, Saba and Havriushenko, Nastia and Herreros, Quentin and G{\"u}nther, Michael and Werk, Maximilian and Xiao, Han},
  journal={arXiv preprint arXiv:2602.15547},
  year={2026}
}

@article{li2026qwen3,
  title={Qwen3-VL-Embedding and Qwen3-VL-Reranker: A Unified Framework for State-of-the-Art Multimodal Retrieval and Ranking},
  author={Li, Mingxin and Zhang, Yanzhao and Long, Dingkun and Chen, Keqin and Song, Sibo and Bai, Shuai and Yang, Zhibo and Xie, Pengjun and Yang, An and Liu, Dayiheng and others},
  journal={arXiv preprint arXiv:2601.04720},
  year={2026}
}

@article{zhang2026f2llm,
  title={{F2LLM}-v2: Inclusive, Performant, and Efficient Embeddings for a Multilingual World},
  author={Zhang, Ziyin and Liao, Zihan and Yu, Hang and Di, Peng and Wang, Rui},
  journal={arXiv preprint arXiv:2603.19223},
  year={2026}
}

@misc{jiang2023mistral7b,
      title={Mistral 7B}, 
      author={Albert Q. Jiang and Alexandre Sablayrolles and Arthur Mensch and Chris Bamford and Devendra Singh Chaplot and Diego de las Casas and Florian Bressand and Gianna Lengyel and Guillaume Lample and Lucile Saulnier and Lélio Renard Lavaud and Marie-Anne Lachaux and Pierre Stock and Teven Le Scao and Thibaut Lavril and Thomas Wang and Timothée Lacroix and William El Sayed},
      year={2023},
      eprint={2310.06825},
      archivePrefix={arXiv},
      primaryClass={cs.CL},
      url={https://arxiv.org/abs/2310.06825}, 
}

@article{grattafiori2024llama,
  title={The llama 3 herd of models},
  author={Grattafiori, Aaron and Dubey, Abhimanyu and Jauhri, Abhinav and Pandey, Abhinav and Kadian, Abhishek and Al-Dahle, Ahmad and Letman, Aiesha and Mathur, Akhil and Schelten, Alan and Vaughan, Alex and others},
  journal={arXiv preprint arXiv:2407.21783},
  year={2024}
}

@article{wilson23koopman,
  title={Koopman operator inspired nonlinear system identification},
  author={D. Wilson},
  journal={SIAM Journal on Applied Dynamical Systems},
  volume={22},
  number={2},
  pages={1445--1471},
  year={2023},
  publisher={SIAM}
}
\vspace{-0.5cm}
\begin{IEEEbiographynophoto}{Dan Wilson}
received the Ph.D. degree in Mechanical Engineering from the University of California, Santa Barbara and is currently an associate professor with the Department of Electrical Engineering and Computer Science at the University of Tennessee, Knoxville, TN, USA. His research focuses on dynamical systems and control theory with a focus on data-driven model identification and reduced order representation of nonlinear dynamical systems. 
\end{IEEEbiographynophoto}
\begin{IEEEbiographynophoto}{Mohamed Akrout}
received the Ph.D. degree in Electrical and Computer Engineering from the University of Manitoba and is currently an assistant professor with the Department of Electrical Engineering and Computer Science at the University of Tennessee, Knoxville, TN, USA. His research interests include statistical signal processing, machine learning, and wireless sensing.
\end{IEEEbiographynophoto}

\vfill

\end{document}